\documentclass[sigconf]{acmart}
\usepackage{booktabs} % For formal tables

\usepackage{multirow}
\usepackage{colortbl}

\setcopyright{rightsretained}
\acmDOI{}

\acmISBN{}

\acmConference[ICVGIP'24]{15th Indian Conference on Computer Vision, Graphics and Image Processing}{December 2024}{Bangalore, India}
\acmYear{2024}
\copyrightyear{2024}

\acmPrice{15.00}

\begin{document}
\title{Graph-Based Multi-Modal Sensor Fusion for Autonomous Driving}
% \titlenote{Produces the permission block, and
  % copyright information}

\author{Depanshu Sani and Saket Anand}
\affiliation{%
  \institution{Indraprastha Institute of Information Technology}
  % \streetaddress{XYZ}
  % \city{XYZ}
  \state{Delhi}
  \country{India}
  % \postcode{000000}
}

\renewcommand{\shortauthors}{}

\begin{abstract}
The growing demand for robust scene understanding in mobile robotics and autonomous driving has highlighted the importance of integrating multiple sensing modalities. By combining data from diverse sensors like cameras and LIDARs, fusion techniques can overcome the limitations of individual sensors, enabling a more complete and accurate perception of the environment. We introduce a novel approach to multi-modal sensor fusion, focusing on developing a graph-based state representation that supports critical decision-making processes in autonomous driving. We present a Sensor-Agnostic Graph-Aware Kalman Filter \cite{saga-kf}, the first online state estimation technique designed to fuse multi-modal graphs derived from noisy multi-sensor data. The estimated graph-based state representations serve as a foundation for advanced applications like Multi-Object Tracking (MOT), offering a comprehensive framework for enhancing the situational awareness and safety of autonomous systems. We validate the effectiveness of our proposed framework through extensive experiments conducted on both synthetic and real-world driving datasets (nuScenes). Our results showcase an improvement in MOTA and a reduction in estimated position errors (MOTP) and identity switches (IDS) for tracked objects using the SAGA-KF. Furthermore, we highlight the capability of such a framework to develop methods that can leverage heterogeneous information (like semantic objects and geometric structures) from various sensing modalities, enabling a more holistic approach to scene understanding and enhancing the safety and effectiveness of autonomous systems.
\end{abstract}




\maketitle

\section{Introduction}
\begin{figure}
    \centering
    \includegraphics[width=0.5\linewidth]{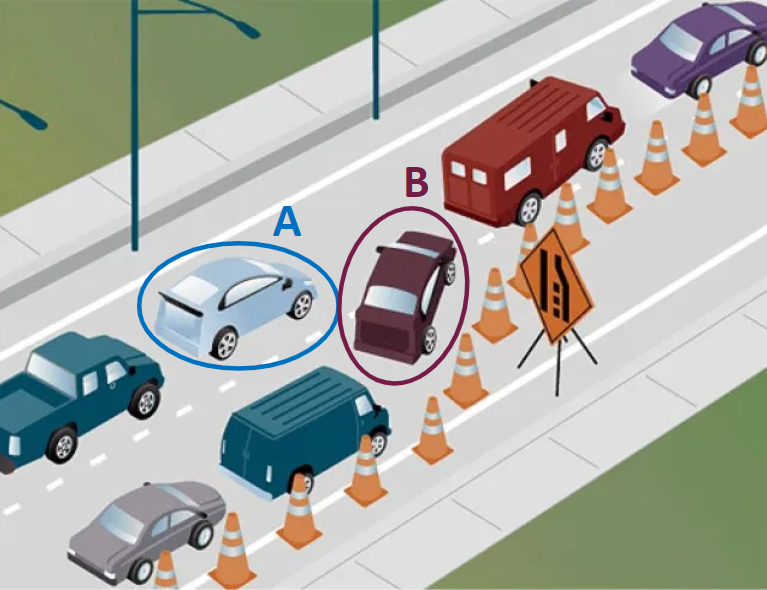}
    \caption{How should car ‘A’ take into account the temporary geometry of the scene formed by cones in order to ensure safety and to avoid any collision with ‘B’?
}
    \label{fig:example}
\end{figure}

In applications of mobile robotics and autonomous driving, the integration of various sensing modalities via multi-modal sensor fusion has become paramount for achieving comprehensive scene understanding that enables effective decision-making. 
% This integration aims to combine the strengths of different sensors to compensate for their individual limitations and provide a more holistic view of the environment. The general methodology for fusing multi-sensor data can be classified into three main categories: early fusion, mid-fusion, and late fusion. Early fusion involves fusing multi-sensor data at the input level of the processing pipeline. Mid-fusion entails extracting and aggregating features from each sensor and inputting the fused features to the predictor. Late fusion involves fusing the predictions obtained by processing each sensor data independently. 
Similar to any other multi-sensor fusion technique, our objective is to leverage the complementarity of the different types of sensors to enable an improved, more holistic view of the environment. For a mobile robot or an autonomous driving (AD) agent, the typical example is that of using a camera and a LIDAR. We present a sensor fusion approach that utilizes cameras and LIDARs mounted on an AD vehicle and aims to build holistic scene representations that facilitate downstream decision-making. Moreover, our proposal also relies on the observation that an AD agent needs both semantic and geometric information about its environment (scene) for decision-making. For instance, in Figure \ref{fig:example}, say the light-blue car (A) in the left lane has a right-of-way at the lane merge ahead. However, the apparent violation of its right-of-way by other agents (B) should be accounted for by its path planner, which in turn would need semantic information, such as the types of vehicle and their behavior in its environment, in addition to the geometric information of the road, lanes and position and velocity of other agents and also the topological information representing the correlation between different entities in the scene. A resulting plan (e.g.,  lowering speed or stopping) that ensures safety needs a holistic understanding of a dynamic environment that can be achieved by effectively processing the multi-modal sensory data to develop appropriate representations that aid decision-making. 

In this extended abstract, we present a \emph{Sensor-Agnostic Graph-Aware Kalman Filter \cite{saga-kf}} and propose to develop a graph-based dynamic scene representation that permits us to capture heterogeneous information from multiple sensing modalities. Using the AD use case, we plan to develop methods for processing dynamic scene graphs that capture semantic (type of vehicle, traffic light, etc.) and geometric (road planes, lane boundary curve, etc.) information as nodes and their pairwise relationships as edges. Our methods will enable inferences drawn using this dynamic scene graph representation with two key applications in AD: Multi-Object Tracking (MOT) and Simultaneous Localization and Mapping (SLAM). 

\begin{figure}
    \centering
    \includegraphics[width=\linewidth]{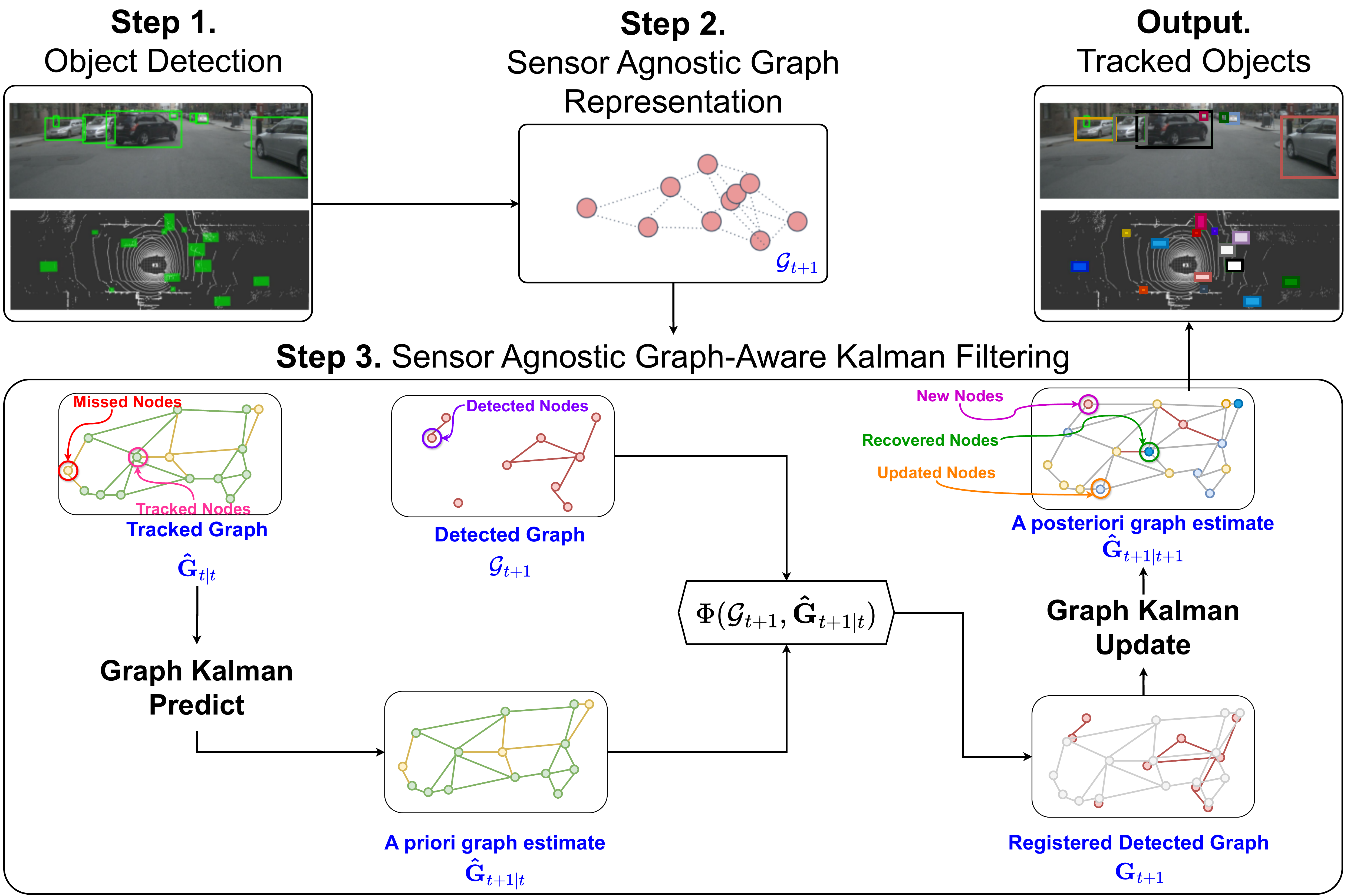}
    \caption{The observed scene graph is constructed using SOTA detectors and is fed into the SAGA-KF framework to predict, associate and estimate the state of the scene graph.}
    \label{fig:enter-label}
\end{figure}

\section{Sensor-Agnostic Graph-Aware Kalman Filter (SAGA-KF)}
Recent progress in open-source object detection techniques has significantly advanced Multi-Object Tracking (MOT) methodologies, primarily under the tracking-by-detection paradigm. To enhance the robustness and reliability of MOT systems, recent research has proposed integrating information gathered from diverse sensors. However, many Kalman filter-based MOT approaches assume the independence of object trajectories, overlooking potential inter-object relationships. 
While some efforts have been made to incorporate these relationships, they often concentrate on learning feature representations to facilitate better association. 

In a recent paper, Bal et al. \cite{aditi-eccv} seek to capture object dependencies using a graph-based representation that explicitly includes inter-object interactions. The objects of interest are represented as nodes of the graph, while interactions between the nodes are captured via edges. Using video as raw input, they formulate MOT as a graph tracking problem, which is solved by designing a Kalman filter over the space of graphs. Tracking a time-series of scene graphs, rather than individual nodes, helps in modeling the dependencies between the constituent objects. 

While \cite{aditi-eccv} clearly establishes the benefits of capturing inter-object relationships using graph representations, its practical applicability and scalability is restricted due to its edge tracking, i.e., estimating mean and covariance of $n^2$ edge attributes. We argue that the inter-object relationships can still be captured in the dynamical model without explicitly tracking graph edges. The resulting focus on node-only tracking can substantially reduce the computational complexity of the Kalman filter. We retain object interactions by imposing a more structured and topology-aware dynamical model on nodes. This model captures inter-node interactions and allows us to share dynamics across the interacting objects. Specifically, we define the time-varying state-transition function as a linear combination of arbitrary but known state-transition functions, called interaction functions $\mathbf{A}^e_t$, that encode the edge information of the tracked graph $\mathbf{\hat{G}}_{t|t}$. As the edges are not explicitly tracked in this approach, our proposed approach is termed graph-aware Kalman filter, as opposed to a graph-tracking Kalman filter. We also emphasize that a graph-based approach allows each object in the scene to be represented as an abstract entity in the scene graph. This abstract representation of the dynamic scene using graph-based representation makes it viable to incorporate different sensing modalities by registering the abstract graphs obtained from each sensor using an assignment method, like the Hungarian algorithm. However, the measurement noise associated with each sensor poses a significant challenge in achieving this. Our dynamical model is also designed to handle the measurement noise associated with each sensor, thereby making the state evolution truly sensor-agnostic. With these two novel components, i.e., a topology-aware node-interacting dynamical model and sensor-agnostic state evolution, we present a Sensor-Agnostic Graph-Aware Kalman Filter (SAGA-KF) and show its efficacy on the MOT problem using a synthetic dataset and the nuScenes \cite{nuscenes} autonomous driving dataset.

\begin{table}
    \caption{Results for the nuScenes dataset on validation set.}
    \label{tab:nuScenes}
    \centering
    \begin{tabular}{|c|c|c|c|c|}
    \hline
        Sensor & Method & AMOTA ($\uparrow$) & AMOTP ($\downarrow$) & IDS ($\downarrow$) \\ \hline
         & \cite{center-point}& \textbf{36.84\%} & \textbf{1.0101} & 870  \\ 
         & Classical KF& 33.57\% & 1.0374 & \textbf{398}  \\ 
        \rowcolor[HTML]{D9EAD3} 
        \multirow{-3}{*}{\cellcolor[HTML]{FFFFFF}Camera} & SAGA-KF & 33.57\% & 1.0375 & 399 \\ \hline

         & \cite{center-point}
& 53.05\% & \textbf{0.7062} & 2521  \\ 
         & Classical KF
& 53.77\% & 0.7422 & \textbf{1363}  \\ 
        \rowcolor[HTML]{D9EAD3} 
        \multirow{-3}{*}{\cellcolor[HTML]{FFFFFF}LiDAR} & SAGA-KF & \textbf{53.82\%} & 0.7418 & 1392 \\ \hline
        
         & \cite{center-point}
& 34.99\% & 0.7626 & 2099  \\ 
         & Classical KF
& 39.71\% & 0.6969 & 960  \\ 
        \rowcolor[HTML]{D9EAD3} 
        \multirow{-3}{*}{\cellcolor[HTML]{FFFFFF}Fusion} & SAGA-KF & \textbf{39.76\%} & \textbf{0.6950} & \textbf{946} \\ \hline   
    \end{tabular}
\end{table}

\section{Proposed Research}
SAGA-KF requires the interaction functions $\mathbf{A}^e_t$, edge connections and edge weights to be known a priori, and hence they are currently hand-crafted for each dataset. Therefore, the resulting estimated graph is sub-optimal. Moreover, our proposed graph-based approach permits the inclusion of heterogeneous nodes generated from multi-modal data, e.g., a node can represent a traffic light in a red state, while another connected neighbor could indicate the stop line for the corresponding traffic light. Tracking these nodes together and their relationship simultaneously is valuable for the path planner of an autonomous vehicle; however, tracking such heterogeneous graphs is challenging as the nodes have different node attributes that are not comparable. To this end, we present some of the future works that we are working on in order to improve the performance and scalability of the proposed framework:
\begin{enumerate}
    \item Design a learning-based technique for modeling complex relationships and the influence of the neighboring nodes. 
    % A natural way to do this is to adopt a Graph Convolutional Network (GCN). The message passing and aggregation mechanism of GCNs can be adopted to learn the interaction functions as well as node representations. We propose to define an appropriate dynamical model using GCNs.
    \item Developing a bipartite graph matching technique that respects the topological structure of the two graphs by imposing domain-specific constraints using edges.
    % Tracking dynamic graphs implicitly involves graph registration that is often solved using the Hungarian or Umeyama algorithms. Instead, we plan to investigate approaches that integrate learned representations that build upon geometric and probabilistic  models, e.g., permutation-invariant GCNs or formulating the assignment problem as an optimization problem in an appropriate RKHS using information theoretic divergences.
    \item Defining an online state estimation method for heterogeneous graphs. Tracking of heterogeneous graphs will be facilitated by the techniques proposed above in 1. and 2.
\end{enumerate}

\bibliographystyle{ACM-Reference-Format}
\bibliography{ICVGIP-Latex-Template}


\begin{thebibliography}{4}

%%% the Web address.

\ifx \showCODEN    \undefined \def \showCODEN     #1{\unskip}     \fi
\ifx \showDOI      \undefined \def \showDOI       #1{#1}\fi
\ifx \showISBNx    \undefined \def \showISBNx     #1{\unskip}     \fi
\ifx \showISBNxiii \undefined \def \showISBNxiii  #1{\unskip}     \fi
\ifx \showISSN     \undefined \def \showISSN      #1{\unskip}     \fi
\ifx \showLCCN     \undefined \def \showLCCN      #1{\unskip}     \fi
\ifx \shownote     \undefined \def \shownote      #1{#1}          \fi
\ifx \showarticletitle \undefined \def \showarticletitle #1{#1}   \fi
\ifx \showURL      \undefined \def \showURL       {\relax}        \fi
% invisible to TeX
\providecommand\bibfield[2]{#2}
\providecommand\bibinfo[2]{#2}
\providecommand\natexlab[1]{#1}
\providecommand\showeprint[2][]{arXiv:#2}

\bibitem[\protect\citeauthoryear{Bal, Mounir, Aakur, Sarkar, and Srivastava}{Bal et~al\mbox{.}}{2022}]%
        {aditi-eccv}
\bibfield{author}{\bibinfo{person}{Aditi~Basu Bal}, \bibinfo{person}{Ramy Mounir}, \bibinfo{person}{Sathyanarayanan Aakur}, \bibinfo{person}{Sudeep Sarkar}, {and} \bibinfo{person}{Anuj Srivastava}.} \bibinfo{year}{2022}\natexlab{}.
\newblock \showarticletitle{Bayesian Tracking of Video Graphs Using Joint Kalman Smoothing and Registration}. In \bibinfo{booktitle}{\emph{ECCV}}. \bibinfo{pages}{440--456}.
\newblock
\showISBNx{978-3-031-19833-5}


\bibitem[\protect\citeauthoryear{Caesar, Bankiti, Lang, Vora, Liong, Xu, Krishnan, Pan, Baldan, and Beijbom}{Caesar et~al\mbox{.}}{2020}]%
        {nuscenes}
\bibfield{author}{\bibinfo{person}{Holger Caesar}, \bibinfo{person}{Varun Bankiti}, \bibinfo{person}{Alex~H. Lang}, \bibinfo{person}{Sourabh Vora}, \bibinfo{person}{Venice~Erin Liong}, \bibinfo{person}{Qiang Xu}, \bibinfo{person}{Anush Krishnan}, \bibinfo{person}{Yu Pan}, \bibinfo{person}{Giancarlo Baldan}, {and} \bibinfo{person}{Oscar Beijbom}.} \bibinfo{year}{2020}\natexlab{}.
\newblock \showarticletitle{nuScenes: A Multimodal Dataset for Autonomous Driving}. In \bibinfo{booktitle}{\emph{CVPR}}.
\newblock


\bibitem[\protect\citeauthoryear{Sani, R~Iyer, Rai, Anand, Srivastava, and Kalyanaraman}{Sani et~al\mbox{.}}{2024}]%
        {saga-kf}
\bibfield{author}{\bibinfo{person}{Depanshu Sani}, \bibinfo{person}{Anirudh R~Iyer}, \bibinfo{person}{Prakhar Rai}, \bibinfo{person}{Saket Anand}, \bibinfo{person}{Anuj Srivastava}, {and} \bibinfo{person}{Kaushik Kalyanaraman}.} \bibinfo{year}{2024}\natexlab{}.
\newblock \showarticletitle{Sensor-Agnostic Graph-Aware Kalman Filter for Multi-Modal Multi-Object Tracking}. In \bibinfo{booktitle}{\emph{Proceedings of the 27th International Conference on Pattern Recognition}}.
\newblock


\bibitem[\protect\citeauthoryear{Yin, Zhou, and Kr{\"a}henb{\"u}hl}{Yin et~al\mbox{.}}{2021}]%
        {center-point}
\bibfield{author}{\bibinfo{person}{Tianwei Yin}, \bibinfo{person}{Xingyi Zhou}, {and} \bibinfo{person}{Philipp Kr{\"a}henb{\"u}hl}.} \bibinfo{year}{2021}\natexlab{}.
\newblock \showarticletitle{Center-based 3D Object Detection and Tracking}.
\newblock \bibinfo{journal}{\emph{CVPR}} (\bibinfo{year}{2021}).
\newblock


\end{thebibliography}

\appendix


% The appendix gets added after the references.


\end{document}